\pdfoutput=1

\documentclass[11pt]{article}

\usepackage{acl}
\usepackage{times}
\usepackage{latexsym}
\usepackage{amsmath}
\usepackage{amssymb}
\usepackage{float}
\usepackage{graphicx}
\usepackage{todonotes}
\usepackage{multirow}
\usepackage{enumitem}

\usepackage[T1]{fontenc}

\usepackage[utf8]{inputenc}

\usepackage{microtype}

\title{Teaching BERT to Wait: Balancing Accuracy and Latency for \\ Streaming Disfluency Detection}

\author{Angelica Chen\Thanks{ Work completed as part of the Student Researcher program at Google.} \\
  New York University \\
  \texttt{ac5968@nyu.edu} \\ \And
  Vicky Zayats \\
  Google Research \\
  \texttt{vzayats@google.com} \\ \And
  Daniel D. Walker \\
  Google Research \\
  \texttt{danwalkeriv@google.com} \\ \AND
  Dirk Padfield \\
  Google Research \\
  \texttt{padfield@google.com} \\
}

\begin{document}
\maketitle
\begin{abstract}
In modern interactive speech-based systems, speech is consumed and transcribed incrementally prior to having disfluencies removed. This post-processing step is crucial for producing clean transcripts and high performance on downstream tasks (e.g. machine translation). However, most current state-of-the-art NLP models such as the Transformer operate non-incrementally, potentially causing unacceptable delays. We propose a streaming BERT-based sequence tagging model that, combined with a novel training objective, is capable of detecting disfluencies in real-time while balancing accuracy and latency. This is accomplished by training the model to decide whether to immediately output a prediction for the current input or to wait for further context. Essentially, the model learns to dynamically size its lookahead window. Our results demonstrate that our model produces comparably accurate predictions and does so sooner than our baselines, with lower flicker. Furthermore, the model attains state-of-the-art latency and stability scores when compared with recent work on incremental disfluency detection.

\end{abstract}

\section{Introduction}

\begin{figure*}[h!]
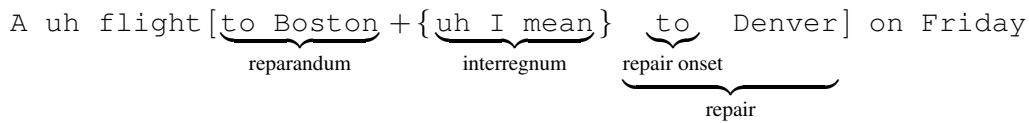

\begin{equation*}
    \,\texttt{A uh flight}\, [\,\underbrace{\texttt{to Boston}}_\text{reparandum} \,+\, \{\,\underbrace{\texttt{uh I mean}}_\text{interregnum}\,\}\, \underbrace{\underbrace{\texttt{to}}_\text{repair onset} \, \texttt{Denver}}_\text{repair} \,] \texttt{ on Friday}
\end{equation*}

\caption{An example of disfluency annotation. \label{fig:disfluency-example}}
\end{figure*}

Many modern Natural Language Understanding (NLU) applications (e.g. transcribers, digital voice assistants, and chatbots) use streaming Automatic Speech Recognition (ASR) systems that incrementally consume speech, offering real-time transcription and predictions with minimal delay. However, these systems are often challenged by the presence of disfluencies, which are unintentional speech disruptions such as ``um'', ``no I meant'', and ``I I I I think,'' that occur naturally in spontaneous speech. Disfluencies not only hurt the readability of ASR transcripts, but also erode model performance on downstream tasks, such as machine translation \cite{hassan2014segmentation} and question answering \cite{gupta-etal-2021-disfl}. Indeed, even state-of-the-art models such as BERT \cite{devlin-etal-2019-bert} and T5 \cite{2020t5} exhibit significant drops in performance (as much as 28 and 20 F1 points, respectively) on the SQuAD-v2 question-answering benchmark \cite{rajpurkar-etal-2018-know} when disfluencies are inserted into the questions \cite{gupta-etal-2021-disfl}. Past work has shown that a prohibitively large amount of data is needed to train an end-to-end dialogue model that is robust to the presence of disfluencies \cite{shalyminov-etal-2017-challenging}. As a result, modern ASR pipelines typically contain a separate post-processing step that detects and removes disfluencies from the transcript, which has been shown to perform better than end-to-end ASR models that generate fluent text from disfluent speech \cite{jamshid-lou-johnson-2020-end}.


\citet{shriberg1997prosody} introduced the following disfluency schema components that are widely used in disfluency detection research: the \emph{reparandum} (spoken segment intended to be removed), the \emph{interruption point} (marked as ``+''), the \emph{repair} (spoken segment that comes as a  replacement to the reparandum, of which the first word is known as the \emph{repair onset}), and the \emph{interregnum} (material that appears between the reparandum and repair). An example of this annotation schema is shown in Figure \ref{fig:disfluency-example}. Usually the disfluency detection task involves identifying and removing the reparandum portion of the disfluency. One of the most popular approaches that targets disfluency detection is the usage of sequence tagging models such as fine-tuned BERT \cite{bach2019noisy,rohanian-hough-2021-best} or LSTM \cite{DBLP:journals/corr/ZayatsOH16,rohanian-hough-2020-framing}.



Another challenge in disfluency detection is the fact that most interactive speech- or text-based applications
consume input incrementally, producing predictions one word at a time, rather than in entire sentences. 
However, recent state-of-the-art pre-trained language models such as BERT have largely been designed for non-incremental processing and are trained only to output predictions on complete input utterances. Using a non-incremental model in an interactive setting produces undesirable delays, since downstream applications must wait for the user to finish their entire utterance before making any decisions.

To address the goal of streaming disfluency detection, recent work has focused on adapting non-incremental models for streaming settings. \citet{madureira-schlangen-2020-incremental} demonstrated that BERT-based models can adequately process incremental input for a variety of sequence tagging tasks when trained on partial sequences, although performance on full sequences suffers. \citet{rohanian-hough-2021-best} applied both the truncated training and prophecy generation strategies from \citep{madureira-schlangen-2020-incremental} to a $\text{BERT}_\text{LARGE}$ model, achieving state-of-the-art performance on streaming metrics among incremental systems. Notably, both these approaches employ the delay strategy of a fixed \emph{lookahead window} - a short amount of right context that the model can ``peek'' at when making its prediction on the current token \cite{buss-schlangen-2011-evaluation}. Although a larger lookahead window can boost accuracy and stability, it also incurs extra delay (by definition).

In the task of incremental disfluency detection, a lookahead window is likely most useful for reparanda, since it is often nearly impossible to identify a reparandum without knowing whether it is followed by an interregnum or repair. However, this extra right context may be much less informative for fluent tokens. Guided by this insight, we extend the past research by training a BERT-based model to dynamically decide how much lookahead context to use. For each new input token that the model consumes, the model can choose to either immediately output a label for that token or to wait for further input before making its prediction. We also design a novel training objective that accounts for both the cross-entropy and latency costs incurred by delaying inference.

In our experiments we explore the trade-offs between accuracy, latency, and output stability for both partial and complete results. To our knowledge, this is the first work to adapt the BERT architecture and training objective to balance accuracy and latency in a streaming sequence tagging task. 

The contributions of our paper are as follows: first, we propose a new model architecture and training objective for streaming sequence tagging tasks. This method involves fine-tuning a pre-trained BERT model to decide when to immediately output predictions and when to wait for further input-- temporarily abstaining from producing a prediction. Secondly, we show that this model achieves high accuracy in incremental settings with state-of-the-art latency and stability, all with a model architecture that is $\sim35$ times smaller than $\text{BERT}_\text{BASE}$ \citep{zhao-etal-2021-extremely}. We demonstrate that the model continues to perform competitively in non-incremental settings when compared to its non-incremental counterparts. Finally, our analyses show that our streaming model learns to wait the most when it encounters an interregnum or reparandum, and the least for fluent or edit terms.

\begin{figure*}[h!]
    \centering
    \includegraphics[width=0.7\textwidth]{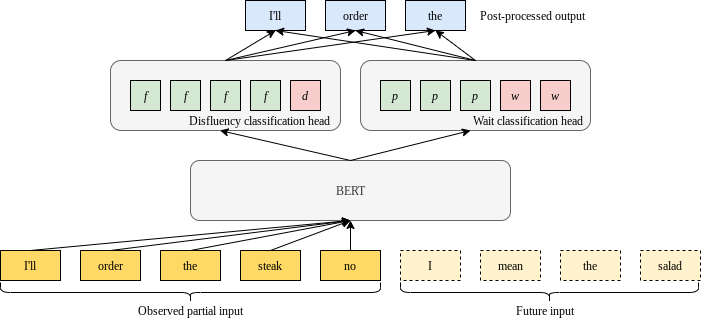}
    \caption{The architecture of our streaming BERT model. The disfluency classification head outputs predictions of \emph{fluent} (\emph{f}) or \emph{disfluent} (\emph{d}) for each token, whereas the wait classification head outputs predictions of \emph{predict} (\emph{p}) or \emph{wait} (\emph{w}) for each token. Given a partial input, we find the first token with a \emph{wait} prediction and output only the tokens before it with \emph{fluent} predictions.}
    \label{fig:model-architecture}
\end{figure*}
\section{Related Work}\label{sec:related-work}
Although disfluency detection itself is a well-studied task, only a handful of past work has explored disfluency detection in an online setting - that is, consuming the input a single token at a time and outputting predictions on the partial input as early as possible. Among the neural approaches, \citet{hough-schlangen-2015-recurrent} were the first to demonstrate competitive performance of recurrent neural networks (RNNs) on incremental disfluency detection by applying an Elman RNN paired with a Markov decoder that jointly optimized the probability of the output tag over the past inputs and outputs. \citet{hough-schlangen-2017-joint} built upon this by jointly training LSTMs on both utterance segmentation and disfluency detection, demonstrating that jointly training on the two tasks yielded higher accuracy and lower latency on both tasks than training on either task alone. This was followed by a number of other works that also successfully paired incremental disfluency detection with other tasks, such as language modeling \citep{shalyminov-etal-2018-multi} and POS tagging \citet{rohanian-hough-2020-framing}.


More recently, large pre-trained transformer architectures have demonstrated incredible success on sequence labeling tasks \citep{vaswani-2017-attention}. Although the original transformer architecture was not designed for streaming input, \citet{chen-2020-controllable} proposed the controllable time-delay transformer instead, which combines a fast decoding strategy with a modified self-attention mechanism that attends only to future inputs in a fixed lookahead window.

The closest work to ours is that of \citet{rohanian-hough-2021-best}, in which the authors fine-tuned a pre-trained $\text{BERT}_\text{LARGE}$ model via add-$M$ training, which feeds the model successive prefixes of lengths $N+M,N+2M,\cdots$ for each full-length training example. Their best-performing model also made use of a prophecy decoding strategy, in which a GPT-2 \citep{radford-2019-language} model predicted the missing right context of each partial input. The BERT model then made its predictions based on the complete extrapolated sequence, POS tags, and word timings. Unlike their work, we aim to train a more lightweight small vocabulary $12\times 128\text{ BERT}$ model that is more suitable for on-device settings, does not require a separate prophecy generation model, and uses only disfluency-annotated training data. We also train our model on successive prefixes of the input (with $N=1,M=1$) but modify both the architecture and training objective in order to balance the competing objectives of accuracy and latency.

\section{Training a Streaming BERT Model}
In this section we describe the architectural changes, new training objective, and training scheme that we use to adapt a (non-incremental) BERT model for streaming sequence tagging tasks. Specifically, we modify the model to enable it to either immediately produce a prediction for a given token or to decide to wait for further input. These changes include a novel training objective that balances the cost between accuracy and latency -- preventing the model from the extremes of either relying too much on waiting for further input or on speedily making predictions at the cost of accuracy. We train this model using a \emph{restart-incremental} training procedure described in Section~\ref{sec:restart_incremental}.

\subsection{Model Design}\label{model-overview}
Streaming settings force models to make trade-offs between accuracy and latency. More accurate predictions can be obtained by providing longer right context at the cost of incurring additional latency. However, since most tokens are not disfluent, a model may not require the right context in order to accurately classify fluent tokens. Rather than using a fixed lookahead window, we train a BERT model to jointly classify tokens and simultaneously choose the lookahead window size dynamically at each token.

Our proposed model architecture consists of a pre-trained BERT model with two separate token classification heads added on top, as shown in Figure \ref{fig:model-architecture}. Each classification head consists of a linear layer applied to the hidden layer outputs of the BERT model. The first classification head, the disfluency classifier, is trained to classify whether each token is disfluent or not. The second classification head (the \emph{wait classifier}) is trained to classify whether the model should wait for further input (temporarily abstain from predicting) or immediately output a prediction for the given token. In effect, the wait classifier decides how large of a lookahead window the model needs to make its prediction on the current input. At inference time, we only output tokens that lie to the left of the first token for which the model outputs a \emph{wait} prediction and that are predicted to be fluent. This avoids potentially producing disjoint output in the case where the model produces predictions of \emph{wait} followed by \emph{predict}, making the output more clear for the user's display.

We also adapt the training objective such that it accounts for both the accuracy and latency of the model's outputs on each successive prefix. Let $(x, y)$ be the pair of input sequence and target output sequence with prefixes $x_1,x_2,\cdots, x_{|x|}$ and $y_1,y_2,\cdots,y_{|x|}$, respectively where $|x|$ is the length of the full sentence. We also denote $f(x)$ as the output logits of the disfluency classifier, $g(x)$ as the output logits of the wait classifier, $\sigma(\cdot)$ as the softmax function, and $H(\cdot,\cdot)$ as the cross-entropy loss. Then the traditional cross-entropy loss on the full input and target sequences is
\begin{equation}
    \ell_\textsc{Full}(x,y) = H(\sigma(f(x)),y)
\end{equation}
However, for each prefix we wish to only compute the cross-entropy loss on the tokens to the left of the first token for which the model outputs a \emph{wait} prediction. To accomplish this, we devise a binary mask that zeros out the loss on the tokens to the right of and including the first \emph{wait} prediction:
\begin{equation}
    \mathbf{m}(\sigma(g(x_i)) = (m_1,\cdots,m_{|x_i|})
\end{equation}
where
\begin{align}
    m_j &= \begin{cases}
                1 & \mbox{if } j < k \\
                0 & \mbox{otherwise}
           \end{cases} \\
    k &= \min\,\{ j \vert \sigma(g(x_i)_j) > 0.5\}
\end{align}
where $g(x_i)_j$ is the $j$-th element of vector $g(x_i)$. We then apply this mask to the cross-entropy loss for each prefix of example $x$ to obtain prefix loss:
\begin{equation}
    \ell_\textsc{Prefix}(x, y) = \sum_{i=1}^{|x|-1} \mathbf{m}(\sigma(g(x_i))) \circ H(\sigma(f(x_i)), y_i)
\end{equation}
where we abuse notation here by denoting $H(\sigma(f(x_i)), y_i)$ as the vector for which the $j$-th element corresponds to the cross-entropy of $(\sigma(f(x_i))_j, y_{i,j})$ and $\circ$ is element-wise multiplication.
Lastly, we define a latency cost that scales with both the probability of abstaining from classifying the $j$-th token in the $i$-th prefix ($\sigma(g(x_{i})_j)$) and with the expected wait time, as measured by number of tokens, incurred by abstaining starting from token $j$ in prefix $x_i$:
\begin{equation}
    \ell_\textsc{Latency}(x)=\sum_{i=1}^{|x|-1}\sum_{j=1}^i (i-j)\sigma(g(x_i)_j)
\end{equation}
Putting these together, the total loss for a single example $(x,y)$ is:
\begin{align}
    \ell(x,y) &= \ell_\textsc{Full}(x,y)+\gamma\ell_\textsc{Prefix}(x,y) \nonumber \\
    &+\lambda\ell_\textsc{Latency}(x)
\end{align}
with hyperparameters $\gamma$ and $\lambda$ controlling the relative strengths of the prefix and latency costs, respectively. We also include the cross-entropy loss on the full sequences ($\ell_\textsc{Full}$) in addition to the prefix losses ($\ell_\textsc{Prefix}$) because we wish for the model to maintain its ability to make predictions on full sequences. Since $g(x)$ does not appear anywhere in $\ell_\textsc{Full}$, the model is effectively forced to make predictions once it receives the full utterance.

Similarly, the $\ell_\textsc{Latency}$ term is essential because without it, the model could achieve minimal loss by always waiting (e.g. $\sigma(g(x_i)_j)=1$ for all prefixes $i$ and time steps $j$), and only learning to classify disfluent tokens after receiving the full sequence. This is equivalent to the non-incremental classification loss. If we instead set $\sigma(g(x_i)_j)=0$ for all $i,j$ (the case where the model never waits), the resulting loss is equivalent to the learning objective for strongly incremental training (see Section \ref{sec:restart_incremental}). In essence, our training objective is a generalization of the strongly incremental objective.

\subsection{Restart-Incremental Training}
\label{sec:restart_incremental}
Although BERT models are typically fine-tuned using complete pre-segmented sequences, an incremental model must process partial inputs at inference time, resulting in a distributional shift between the complete utterances typically seen in training datasets and the partial utterances seen at inference time. A simple solution is to fine-tune BERT both on complete and partial inputs, a training scheme known as \emph{restart incrementality} \citep{kahardipraja-2021-incremental}. By providing successively extended prefixes of a given utterance to the model and computing the loss on the model outputs for each prefix, we can mimic the streaming data that the model would encounter in real time. In all of our experiments, each successive prefix adds a single word to the previous prefix, a setting known as \emph{strongly incremental} \citep{shalyminov-etal-2018-multi}. 
Although this approach requires re-computation of the model outputs for each successive prefix, this also enables the model to correct its previous predictions, or to switch between waiting and predicting when it receives helpful right context. 
Incorporating prefixes during training in incremental disfluency detection has been previously explored by \citet{rohanian-hough-2021-best}. This serves as a strong baseline in our experiments.

\section{Experimental Setup}
We fine-tune all models on the Switchboard dataset \citep{godfrey-1992-switchboard}, a transcribed English multi-speaker conversational corpus that is commonly used for ASR research. We specifically use the version from the Linguistic Data Corpus's Treebank-3 \citep{mitchell-1999-treebank3} distribution, which additionally contains disfluency annotations and a standard train/dev/test split \citep{charniak-johnson-2001-edit}. We follow \citet{rocholl-2021-disfluency}, training our models to classify both the reparanda and  interregna as disfluent for future removal in a final post-processed transcript.

\subsection{Baselines}
All of our experiments use small distilled $\text{BERT}$ models, specifically a small vocabulary $\text{BERT}$ model \citep{zhao-etal-2021-extremely} ($\text{BERT}_\text{SV}$) with 12 hidden layers of size 128 that is pre-trained on English Wikipedia and BookCorpus \citep{zhu-2015-bookcorpus}. The details of our hyperparameter tuning can be found in the Appendix (Section \ref{sec:appendix-hp-tuning}).

We use small models for two reasons: 1) fine-tuning a model on all given prefixes of each training example is resource intensive, and 2) many streaming natural language understanding applications run entirely on mobile devices which precludes the use of large models. Previous work on small non-incremental BERT-based models used for disfluency detection \cite{rocholl-2021-disfluency} showed significant improvement in memory and latency without compromising task performance. The core $\text{BERT}_\text{SV}$ model is a distilled version of $\text{BERT}_\text{BASE}$ with smaller vocabulary and reduced hidden layer dimensions \cite{zhao-etal-2021-extremely}. Due to its smaller vocabulary size (5K versus 30K tokens), the model has only about 3.1M parameters, as compared to $\text{BERT}_\text{BASE}$'s approximately 108.9M parameters, achieving around 80\% latency reduction.

In order to isolate the effects of training with \emph{restart incrementality} (Section~\ref{sec:restart_incremental}) versus the improvements derived directly from incorporating our new training objective, we also evaluate two other models: 1) a non-incremental $\text{BERT}_\text{SV}$ model trained in the usual way, on full sequences; and 2) a $\text{BERT}_\text{SV}$ model trained with \emph{restart incrementality} - i.e., on all prefixes of every training example (which we will refer to as ``all prefixes'' in following tables). Setup (1) is equivalent to ablating both $\ell_\textsc{Prefix}$ and $\ell_\textsc{Latency}$ whereas setup (2) is equivalent to ablating only $\ell_\textsc{Latency}$. We do not ablate $\ell_\textsc{Prefix}$ in isolation since this leaves only the $\ell_\textsc{Full}$ and $\ell_\textsc{Latency}$ terms, and there does not exist a meaningful measure of latency when the model never needs to wait for more input (since it is always given the full utterance as input). For each of these baseline models we also follow \citet{rohanian-hough-2021-best} and \citet{kahardipraja-2021-incremental} by evaluating with different fixed lookahead (LA) window sizes of $\text{LA}=0,1,2$.

\begin{table*}[h!]
\centering
\begin{tabular}{p{0.19\textwidth}|p{0.17\textwidth}|cccccc|c}
\hline
\multirow{2}{*}{\textbf{Model}} & \multirow{2}{*}{\textbf{Training Scheme}} & \multicolumn{6}{c|}{\textbf{Incremental}} & \multicolumn{1}{c}{\textbf{Final}} \\
 & & $F_1\uparrow$ & $P\uparrow$ & $R\uparrow$ & EO $\downarrow$ & TTD $\downarrow$ & AWT $\downarrow$ & $F_1\uparrow$ \\ \hline
$\text{BERT}_\text{SV}$           & Full sequences & 0.76 & 0.74 & \textbf{0.78} & 0.31 & 1.46 & \textbf{0.00} & \textbf{0.89} \\
$\text{BERT}_\text{SV}$         & All prefixes &  0.76 & 0.73 & \textbf{0.78} & 0.32 & \textbf{1.37} & \textbf{0.00} & \textbf{0.89} \\
Streaming $\text{BERT}_\text{SV}$    & All prefixes  & \textbf{0.83} & \textbf{0.92} & 0.75 & \textbf{0.09} & 2.32 & 0.21 & 0.88 \\
\hline
\multicolumn{9}{l}{Models with lookahead $\geq 1$ } \\ \hline
$\text{BERT}_\text{SV}$ ($\text{LA}=1$) & Full sequences & 0.83 & 0.85 & 0.80 & 0.10 & 2.41 & 1.00 & 0.89  \\
$\text{BERT}_\text{SV}$ ($\text{LA}=2$) & Full sequences & 0.85 & 0.89 & 0.82 & 0.05 & 3.06 & 2.00 & 0.89 \\
$\text{BERT}_\text{SV}$ ($\text{LA}=1$) & All prefixes & 0.82 & 0.85 & 0.80 & 0.12 & 2.33 & 1.00 & 0.89 \\
$\text{BERT}_\text{SV}$ ($\text{LA}=2$) & All prefixes & 0.85 & 0.89 & 0.82 & 0.06 & 3.01 & 2.00 & 0.89 \\ \hline
\end{tabular}
\caption{Comparison of incremental performance on the Switchboard validation set of non-incremental small-vocab BERT models ($\text{BERT}_\text{SV}$) against that of a streaming small-vocab BERT model (streaming $\text{BERT}_\text{SV}$). In the lower half of the table we also list the evaluation results of non-incremental $\text{BERT}_\text{SV}$ models with fixed lookahead (LA) window sizes of 1 and 2 tokens. Note that for the non-incremental models the lookahead window size is equivalent to the average waiting time (AWT). The arrows near each metric represent the desirable direction of the result: $\uparrow$ means the higher the performance the better and $\downarrow$ is the reverse.}
\label{tab:streaming-eval}
\end{table*}
\subsection{Incremental Evaluation}
Accuracy alone is not a sufficient measure of success to robustly evaluate a streaming model. Since a streaming model is meant to operate in real time, it should return output as soon as possible after it receives new input. As such, we also need to evaluate it with respect to latency -- i.e. the number of new tokens a model must consume before producing a prediction for the current token. Furthermore, streaming models are often designed to be capable of retroactively changing their predictions on previous tokens as new input arrives. This introduces the risk of output ``jitter'' or ``flicker,'' where the output changes dramatically as new input is consumed, necessitating evaluation of stability. To capture all these important dimensions of streaming model performance, we evaluate the models using the following diachronic metrics (with formulas and further details in the \hyperref[sec:appendix-eval-metrics]{Appendix}):
\begin{itemize}
    \item \textbf{Streaming $F_1$}: An accuracy metric scored in the same way as the typical $F_1$ score, albeit we score the predictions for a single token over the course of multiple time steps separately as if they were predictions for separate tokens.
    \item \textbf{Edit Overhead (EO)} \citep{buss-schlangen-2011-evaluation}: A stability metric that measures the average number of unnecessary edits, normalized by utterance length.
    \item \textbf{Time-to-detection (TTD)} \citep{hough-schlangen-2017-joint}: A latency metric that is only computed on disfluent tokens that are classified correctly. It is the average amount of time (in number of tokens consumed) that the model requires before first detecting a disfluency. As mentioned earlier, we include both reparanda and interregna as disfluencies. 
    \item \textbf{Average waiting time (AWT)}: The average amount of time (in number of tokens consumed) that the model waits for further input before making a prediction on a given token. For models with a fixed lookahead window, this is equivalent to the lookahead window size. For the streaming model, this is equivalent to the average lookahead window size.
    \item \textbf{First time to detection (FTD)} \citep{zwarts-etal-2010-detecting, rohanian-hough-2021-best}:  Similar to to the TDD metric described above with the main difference being that the latency (in number of words) is calculated starting from the onset of a gold standard repair.
\end{itemize}

\section{Results}
In this section we present a summary of both the non-incremental and incremental performance of our streaming model against that the baselines. We also present an analysis of the types of errors and average amount of waiting time the streaming model incurs.

\subsection{Incremental Performance}
Table \ref{tab:streaming-eval} shows both the incremental and non-incremental evaluation metrics. Our proposed streaming $\text{BERT}_\text{SV}$ model achieved a $9\%$ increase\footnote{All percentages mentioned in this section are computed as a percentage of the original number, rather than as a difference in percentage points.} in streaming $F_1$ over both of the baselines (with lookahead = 0), as well as a $71\%$ and $72\%$ decrease in edit overhead compared to the non-streaming models trained on full sequences and all prefixes, respectively. Despite being trained with a different architecture and loss objective, the streaming model does not sacrifice its non-incremental performance, yielding a final output $F_1$ score that is only one point less than its non-streaming counterparts. Generally speaking, when the streaming model does output a prediction, it classifies tokens as disfluent less often than the non-streaming models with zero LA window, achieving much higher precision (P) and marginally lower recall (R), resulting in a model that ``flickers" less frequently. However, this does contribute to a slightly higher time-to-detection score compared to the baselines with zero lookahead, since the streaming model is generally less aggressive but more precise with outputting disfluent predictions. When compared to the models with fixed lookahead (the lower half of Table \ref{tab:streaming-eval}), however, the streaming model always achieves lower TTD while achieving significantly lower waiting time and comparable accuracy and stability.

\begin{table}[h!]
    \centering
    \begin{tabular}{l|c}
        \textbf{Type of disfluency} & \textbf{Average wait time} \\ \hline
        Repair & 0.74 \\
        Fluent & 0.15 \\
        Interregnum & 1.06 \\
        Reparandum & 0.76 \\
        Edit & 0.14 \\
        Repair onset & 0.46 \\\hline
    \end{tabular}
    \caption{The streaming model's average waiting time (in number of tokens) for each type of token (as categorized in \citet{mitchell-1999-treebank3}) encountered in each prefix fed to the model for the Switchboard validation set. For a more fine-grained analysis, we separate the repair onset (the first word in the repair phrase) from the rest of the words in the repair. The category ``Edit'' consists of all edit terms that are not interregna (i.e. not inside of a repair structure). \label{tab:wait-time-by-disfluency-type}}
\end{table}

\subparagraph{Effect of lookahead window size} We also evaluated the performance of the non-streaming baseline models with fixed lookahead window sizes of 1 and 2 tokens, as shown in the lower half of Table \ref{tab:streaming-eval}. In line with what has been reported in past work \citep{madureira-schlangen-2020-incremental, buss-schlangen-2011-evaluation}, the size of the lookahead window scales directly with the accuracy and stability and inversely with the latency of the model. However, the streaming model has comparable streaming $F_1$ and edit overhead scores as the non-streaming models with $LA=1$, even though the streaming model has 79\% less average wait time. This indicates that the streaming model is able to correctly classify tokens sooner and with more stability than the baseline models that have $LA=1$. Although the models with $LA=2$ improve marginally on accuracy and stability over the models with $LA=1$, the streaming model continues to have lower TTD and AWT but comparable $F_1$ and EO when compared to the models with $LA=2$. 

\subparagraph{The utility of dynamic lookahead} The results in Table \ref{tab:streaming-eval} also reveal some insights into which parts of the model design and training scheme are more important for streaming performance and efficiency. Merely training a non-streaming model on prefixes of the training examples appears to have minimal effect on $F_1$, precision, and recall, but does somewhat improve the TTD score. We hypothesize that this is largely the result of training on a data distribution that more closely resembles the test distribution. Adding the extra wait classifier head and latency cost term in the training objective yields the greatest improvements in both precision and stability, as seen in the differences in $F_1$, P, and EO values between the $\text{BERT}_\text{SV}$ model trained on all prefixes and the streaming $\text{BERT}_\text{SV}$ model.

\begin{table*}[h!]
\centering
\begin{tabular}{ll|cc}
\hline
\textbf{Model} & \textbf{Training Scheme} & \multicolumn{2}{c}{\textbf{Incremental metrics}} \\
 & &  \textbf{EO}$\downarrow$ & \textbf{FTD}$\downarrow$ \\
\hline
$\text{BERT}_\text{LARGE}$  \citep{rohanian-hough-2021-best} & All prefixes             & 0.60 & 0.31 \\ 
\hline
$\text{BERT}_\text{SV}$          & Full sequences       & 0.30 & 0.79 \\
$\text{BERT}_\text{SV}$          & All prefixes         & 0.32 & 0.84 \\
Streaming $\text{BERT}_\text{SV}$ & All prefixes         & \textbf{0.09}  & \textbf{0.11} \\
\hline
\end{tabular}
\caption{\label{competitor-results}
A comparison of the EO and FTD metrics of our baselines ($\text{BERT}_\text{SV}$ trained on full sequences and all prefixes), our streaming $\text{BERT}_\text{SV}$ model, and \citet{rohanian-hough-2021-best}'s incrementalized $\text{BERT}_\text{LARGE}$ model on the Switchboard test set. The arrows near each metric represent the desirable direction of the result -- for both of the metrics, lower numbers are more desirable.
}
\end{table*}
\subparagraph{When to wait} Since the streaming model can abstain from outputting predictions for arbitrarily long suffixes of the input, it incurs waiting time - an average of 0.21 tokens more than the non-streaming models with 0 lookahead. Table \ref{tab:wait-time-by-disfluency-type} shows that the streaming model abstains the most when encountering interregna and reparanda, waiting for approximately 1.06 and 0.76 more tokens, respectively. Given that it is easier to identify a disfluency once the entire reparandum and interregnum have been observed, it follows that the model's predictions may be more uncertain for reparanda and interregna upon first consumption, thus incurring the highest average waiting times. An example of the model's incremental outputs for a disfluency structure is shown in Table \ref{tab:model-outputs-example}. For correctly classified disfluent tokens, the streaming model also has a higher TTD, likely because the non-incremental models are more aggressive in predicting disfluent labels (while making more errors) than the streaming model. Still, this TTD is lower than all the models with fixed lookahead windows.

\subsection{Error Analysis}
\begin{figure}[h!]
    \centering
    \includegraphics[width=0.5\textwidth]{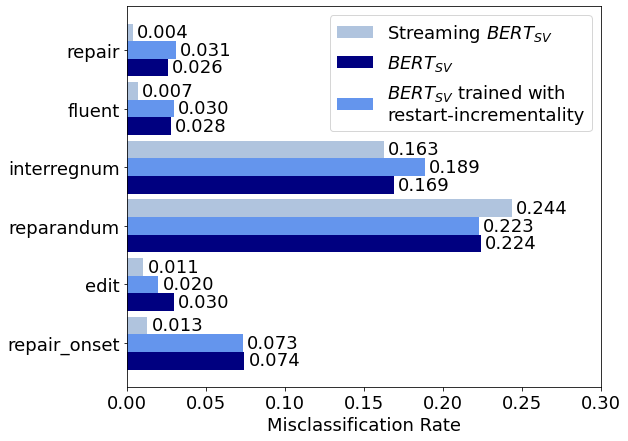}
    \caption{Token-wise misclassification rates for each type of disfluency annotation (as categorized in \citet{mitchell-1999-treebank3}) across all three evaluated models on the Switchboard validation set. If the model abstained from making a prediction on a particular token, we did not count this as an error.}
    \label{fig:wer}
\end{figure}

Figure \ref{fig:wer} shows an error analysis on the models' predictions. We computed the percentage of the time that the streaming model misclassified a token, counting each incidence of the token across each time step separately. All models achieved the lowest misclassification rates on fluent and edit tokens, and the highest misclassification rates on reparanda tokens. For tokens that were fluent, edit terms, or part of a repair or repair onset, the streaming model achieved significantly lower misclassification rates than the baselines. However, all three models performed comparably on interregna and reparanda. Since we measure misclassification rate on every token at every time step, the high misclassification rates on reparanda are expected as it is often not feasible to detect a reparandum until an interregnum or repair onset has been seen.

\subparagraph{Accuracy versus latency} 
In comparison to the baseline models with 0 lookahead, the streaming model makes the largest tradeoffs in accuracy versus latency for repair onsets and repairs, as shown by Figure \ref{fig:wer} and Table \ref{tab:wait-time-by-disfluency-type}. While the streaming model incurs average additional wait times of 0.74 and 0.46 tokens for repairs and repair onsets respectively, its misclassification rates are also approximately $85\%$ and $82\%$ less than the baseline models on repairs and repair onsets respectively. In addition, Table \ref{tab:streaming-eval} demonstrates that the streaming model still achieves comparable $F_1$ and greater stability (lower EO) in comparison to the non-streaming baselines with lookahead 1, despite having an average wait time that is $79\%$ shorter.

\subsection{Comparison with Competitor Baselines}
 As shown in Table \ref{competitor-results}, in comparison with the $\text{BERT}_\text{LARGE}$-based prophecy decoding model proposed in \citet{rohanian-hough-2021-best}, our streaming model achieves state-of-the-art stability (85\% decrease in EO) and latency (65\% decrease in FTD)\footnote{In addition to shorter word-level latency metrics presented in the results, the runtime latency of the $\text{BERT}_\text{SV}$ model is 80\% lower than that of $\text{BERT}_\text{BASE}$ \cite{rocholl-2021-disfluency}.}, despite having far fewer parameters.
\section{Conclusion and Future Work}

We have introduced a streaming BERT-based Transformer model that is capable of balancing accuracy with latency by simultaneously making token-level disfluency predictions and dynamically deciding how large of a lookahead window to use. Our approach improves both streaming accuracy and output stability on an incremental disfluency detection task. Furthermore, it incurs very low average latency in comparison with non-incremental BERT models of the same size. Lastly, our model requires minimal lookahead beyond disfluent regions and achieves state-of-the-art edit overhead and first-time-to-detection scores compared to past work \citep{rohanian-hough-2021-best}. 



While the main focus of this paper has been on developing a fast, accurate, and stable streaming model for disfluency detection, our approach is general enough to be used in other incremental tagging models of linguistic phenomena that benefit from the right context for optimal accuracy. In future work we are interested in applying this approach to tasks such as real-time punctuation prediction and incremental parsing.

Furthermore, the streaming model's efficiency is limited by its non-autoregressive nature and training via restart incrementality. Future work should also explore how to apply a dynamic lookahead window without re-computing the predictions on all the previous inputs and to rely on fewer prefixes during training.

\section{Ethical Considerations}
While our work does not introduce a new dataset, it does depend on a training dataset that was collected from fluent English-speaking, able-bodied human subjects. If deployed in a real-world application, this model would likely perform noticeably worse for users who speak with non-American accents or speech impediments. Transcripts for these users could be disproportionately noisy and the streaming model's average wait time would likely also be longer. Care should be taken to assess the sensitivity and robustness of such a model to non-fluent or non-American English prior to deployment. This model should also be used very cautiously in situations where mistakenly eliding fluent portions of speech from the captions or transcript could incur dire consequences, such as in an emergency call center.

\section{Acknowledgments}
We thank Johann Rocholl, Colin Cherry, Dan Liebling, Noah Murad, and members of the Google Research Speech Intelligence team for valuable feedback and discussion. We also thank the Google Student Researcher and research internship programs for supporting this work. Lastly, we also thank the anonymous reviewers for their thorough and helpful comments.
\bibliography{anthology,custom}
\bibliographystyle{acl_natbib}

\appendix
\section{Appendix}\label{sec:appendix}
\subsection{Evaluation Metrics}\label{sec:appendix-eval-metrics}
We provide more detailed formulas here for each of our evaluation metrics. For a given input utterance $x$, let $x[j]$ be the $j$-th token of $x$, $y[j]$ be the gold label (either 0 for fluent or 1 for disfluent) for $x[j]$, $x[:i]$ be the $i$-th prefix of $x$ (i.e. the first $i$ tokens of $x$), $f(x[:i])[j]$ be the predicted label for $x[j]$ after the model has consumed prefix $i$ , $D$ be the entire dataset that we are evaluating predictive performance for, and $|D|$ be the size of $D$. 

\paragraph{Accuracy metrics} Our definitions of the streaming true positives, true negatives, false positives, and false negatives are:
\begin{align}
    TP_\text{streaming} &= \left|\{(i,j) | y[j]=1,f(x[:i])[j]=1\}\right| \nonumber \\
    TN_\text{streaming} &= \left|\{(i,j) | y[j]=0,f(x[:i])[j]=0\}\right| \nonumber \\
    FP_\text{streaming} &= \left|\{(i,j) | y[j]=0,f(x[:i])[j]=1\}\right| \nonumber \\
    FN_\text{streaming} &= \left|\{(i,j) | y[j]=1,f(x[:i])[j]=0\}\right| \nonumber 
\end{align}
Similar to the traditional definitions, streaming precision, recall, and $F_1$ are computed as:
\begin{align}
    P_\text{streaming} &= \frac{TP_\text{streaming}}{TP_\text{streaming}+FP_\text{streaming}} \nonumber \\
    R_\text{streaming} &= \frac{TP_\text{streaming}}{TP_\text{streaming}+FN_\text{streaming}} \nonumber \\
    F1_\text{streaming} &= 2 \times \frac{P_\text{streaming}\times R_\text{streaming}}{P_\text{streaming} + R_\text{streaming}} \nonumber
\end{align}

\paragraph{Stability metrics}
\begin{itemize}
\item \textbf{Edit overhead (EO)} \citep{buss-schlangen-2011-evaluation}: To calculate edit overhead, we need to first identify, for each prefix $x[:i]$ of a given input $x$, which tokens $x[j]$ have predictions $f(x[:i])[j]$ that differ from the model's prediction in the previous prefix $f(x[:i-1])[j]$. Denoting the cardinality of this set as $E(x)$ (for the number of edits the model makes on $x$), we have:
\begin{equation*}
    E(x)=|\{(i,j)| f(x[:i])[j] \neq f(x[:i-1])[j]\}|
\end{equation*}
Then we can compute EO as follows:
\begin{equation*}
    EO = \frac{1}{|D|}\sum_{x\in D} \frac{E(x)}{|x|} 
\end{equation*}
where $|x|$ is the number of tokens in $x$.
\end{itemize}

\paragraph{Latency metrics}
\begin{itemize}
\item \textbf{Time-to-detection (TTD)} \citep{hough-schlangen-2017-joint}: Since time-to-detection (TTD) is measured only on disfluent tokens that are eventually predicted as such by the model, we need to first define the set of tokens  in a given example $x$ that are true positives at some point:
\begin{equation*}
    TP(x) = \{x[k] | y[k]=1, \exists i:f(x[:i])[k]=1 \}
\end{equation*}
Then for a given token $x[k]\in x$, the detection time (DT) can be calculated as:
\begin{equation*}
    DT(x[k]) = \min_i\,\{i|f(x[:i])[k]=1\}-k
\end{equation*}
It follows that the TTD for the entire dataset $D$ is the average $DT$ for all disfluent tokens that are eventually detected for all $x\in D$:
\begin{equation*}
    TTD = \frac{1}{m}\sum_{x[k]\in TP(x), x\in D} DT(x[k])
\end{equation*}
where
\begin{equation*}
    m = |D|\sum_{x\in D}|TP(x)|,
\end{equation*}
the total number of disfluent tokens in the dataset that are eventually detected by the model.

\item \textbf{First time to detection (FTD)} \citep{zwarts-etal-2010-detecting, rohanian-hough-2021-best}: Similar to TTD, this metric is only measured on disfluent tokens that are eventually detected by the model. Then given some $x[k]\in TP(x)$, let $RI(x[k])$ represent the index of the first token in the repair that follows $x[k]$. Since we are measuring detection time from the start of a gold standard repair instead, the detection time becomes:
\begin{align*}
    DT(x[k]) &= \min_i\,\{i|f(x[:i])[k]=1\} \\
             &- RI(x[k])
\end{align*}
The rest of the formula for the FTD is similar to that of the TTD:
\begin{equation*}
    FTD = \frac{1}{m}\sum_{x[k]\in TP(x), x\in D} DT(x[k])
\end{equation*}
where
\begin{equation*}
    m = |D|\sum_{x\in D}|TP(x)|,
\end{equation*}
the total number of disfluent tokens in the dataset that are eventually detected by the model.

\item \textbf{Average waiting time (AWT)}: Suppose that given an input token $x[k]$, the model can abstain from making a prediction (which occurs both with the streaming model and with the fixed lookahead models). We denote this outcome as $y[k]=\varnothing$. To compute AWT, we first calculate the first prediction time (FPT) for a given token $x[k]$,
\begin{equation*}
    FPT(x[k]) = \arg\min_i \{i|f(x[:i])[k] \neq \varnothing \},
\end{equation*}
i.e. the first time step $i$ in which the model outputs a prediction for token $x[k]$. Then the AWT is
\begin{equation*}
    AWT = \frac{1}{m}\sum_{x[k]\in x, x\in D}FPT(x[k])-k,
\end{equation*}
where
\begin{equation*}
    m=|D|\sum_{x\in D}|x|,
\end{equation*}
the total number of tokens in the dataset.

\end{itemize}

\subsection{Model Training and Hyperparameter Tuning}\label{sec:appendix-hp-tuning}
We implemented our models using TensorFlow v2.7 \citep{tensorflow2015-whitepaper} and the Hugging Face \texttt{transformers} library \citep{wolf-etal-2020-transformers}. We also fine-tuned all model hyperparameters using Vizier \citep{golovin-vizier-2017}, a black-box optimization system, using streaming F1 score on the Switchboard validation set as our objective. The searched ranges for each hyperparameter were $\text{learning rate}\in [1\times 10^{-5}, 1\times 10^{-1}]$, $\text{number of training epochs}\in [12, 20]$, $\lambda\in [1\times 10^{-8}, 1\times 10^{-6}]$, $\gamma\in [1, 10]$. For most experiments we ran 30 trials total, with 10 evaluations in parallel. Each individual trial (one set of hyper-parameters) ran on a single NVIDIA P100 GPU. Experimental run time varied from about 13 to 24 hours, depending mostly on the number of epochs. For each model variant we present only the results from the configuration with the highest streaming $F_1$ score on the Switchboard validation dataset. Our best performing streaming model used parameter values of $\lambda=1.5\times 10^{-7}$, learning rate $1.2\times 10^{-4}$, $\gamma=1.9$, training batch size 8, and 12 epochs.

\begin{table}[t!]
\centering
\begin{tabular}{p{0.05\textwidth}|p{0.37\textwidth}}
    \textbf{Time step} & \textbf{Model outputs} \\ \hline
    \multirow{2}{*}{3} & \textbf{Input}: ``I think [the real,''\\
      & \textbf{Output}: ``I think the real'' \\ \hline
    \multirow{2}{*}{4} & \textbf{Input}: ``I think [the real, + the''\\
      & \textbf{Output}: ``I think the $\texttt{<WAIT>}$'' \\ \hline
    \multirow{2}{*}{5} & \textbf{Input}: ``I think [the real, + the principal]''\\
      & \textbf{Output}: ``I think $\texttt{<DIS>}$ $\texttt{<DIS>}$ the principal'' \\ \hline
\end{tabular}
\caption{Example of the model's outputs at each time step. (For brevity, we excerpt only a segment of the sentence that contains disfluencies.) A $\texttt{<WAIT>}$ symbol indicates that the model decided to stop making predictions for the rest of the input sequence and to wait for further input instead. A $\texttt{<DIS>}$ symbol indicates that the corresponding input token was given a classification of \emph{disfluent} and therefore not included in the final edited output. For clarity we provide disfluency annotations in the form $\texttt{[Reparandum, + Repair]}$, but these are not actually provided to the model as input. \label{tab:model-outputs-example}}
\end{table}



\end{document}